\documentclass[10pt, twocolumn]{IEEEtran}

\usepackage{url}
\usepackage{comment}
\usepackage[ruled]{algorithm2e}

\SetAlFnt{\small}
\SetAlCapFnt{\small}

\usepackage{amssymb,amsfonts,amsmath}
\usepackage{graphicx}
\usepackage{algorithmic}
\usepackage{color}

\newtheorem{theorem}{Theorem}
\newtheorem{corollary}[theorem]{Corollary}
\newtheorem{lemma}{Lemma}
\newtheorem{definition}{Definition}

\def\va{ {\vec{a}} }
\def\pv{ {\vec{p}} }

\def\vx{ {\vec{x}} }
\def\N{ {\mathbb{N} }}

\def\Dc{ {\mathcal{D} }}

\def\Nc{ {\mathcal{N} }}

\def\Mc{ {\mathcal{M} }}

\def\nox{(D3) }
\def\noM{(D4) }
\def\noPhi{(D5) }
\def\findsoln{(D1) }
\def\sticksoln{(D2) }

\begin{document}

\title{Decentralized Constraint Satisfaction}

\author{K. R. Duffy$^1$, C. Bordenave$^2$ and D. J. Leith$^1$.\\
$1$: Hamilton Institute, National University of Ireland Maynooth\\
$2$: Department of Mathematics, University of Toulouse.\thanks{This material is based upon works supported by Science Foundation
Ireland under Grant No. 07/IN.1/I901.
}}

\maketitle

\begin{abstract}
We show that several important resource allocation problems in
wireless networks fit within the common framework of Constraint
Satisfaction Problems (CSPs). Inspired by the requirements of these
applications, where variables are located at distinct network devices
that may not be able to communicate but may interfere, we define
natural criteria that a CSP solver must possess in order to be
practical. We term these algorithms decentralized CSP solvers. The
best known CSP solvers were designed for centralized problems and
do not meet these criteria. We introduce a stochastic decentralized
CSP solver, proving that it will find a solution in almost surely
finite time, should one exist, and also showing it has many practically
desirable properties. We benchmark the algorithm's performance on
a well-studied class of CSPs, random k-SAT, illustrating that the
time the algorithm takes to find a satisfying assignment is competitive
with stochastic centralized solvers on problems with order a thousand
variables despite its decentralized nature. We demonstrate the
solver's practical utility for the problems that motivated its
introduction by using it to find a non-interfering channel allocation
for a network formed from data from downtown Manhattan.

\end{abstract}

\section{Introduction} 
\label{sec:intro}

A Constraint Satisfaction Problem (CSP) consists of $N$ variables,
$\vx:=(x_1,\ldots,x_N)$, and $M$ clauses, i.e. $\{0,1\}$-valued
functions, $(\Phi_1(\vx),\ldots,\Phi_M(\vx))$. An assignment $\vx$
is a solution if all clauses simultaneously evaluate to $1$. In the
context of wireless networks, we show that CSPs provide a unifying
framework that encompasses many important resource allocation tasks.
Examples include: allocation of radio channels so that transmissions
by neighbouring WLANs (Wireless Local Area Networks) or mobile
phone cells do not interfere; the selection of packets to be XORed
on each link in network coding; and finding a non-colliding schedule
of time-slots in a WLAN. Unlike in traditional CSPs, however, in
network applications each constrained variable $x_i$ is typically
co-located with a physically distinct device such as an access
point, base-station or a link. The resulting communication constraints
impose severe restrictions on the nature of the algorithm that can
be used for solving the CSP. These restrictions are violated by
existing CSP solvers, leading us to define a new class of algorithms
that we term \emph{decentralized CSP solvers}. We constructively
establish the existence of decentralized solvers by introducing a
family of randomized algorithms belonging to this class.

Roughly speaking, decentralized CSP solvers must be capable of
finding a satisfying assignment, $\vx$, while updating each variable
$x_i$ based solely on knowledge of whether all of the clauses in
which $x_i$ participates are satisfied or at least one clause is
unsatisfied. The following example gives a concrete illustration
of the origins of these constraints.

\subsection{Motivating example: channel allocation}
\label{sec:channel_all}

\begin{figure}
\centering
\includegraphics[width=0.8\columnwidth]{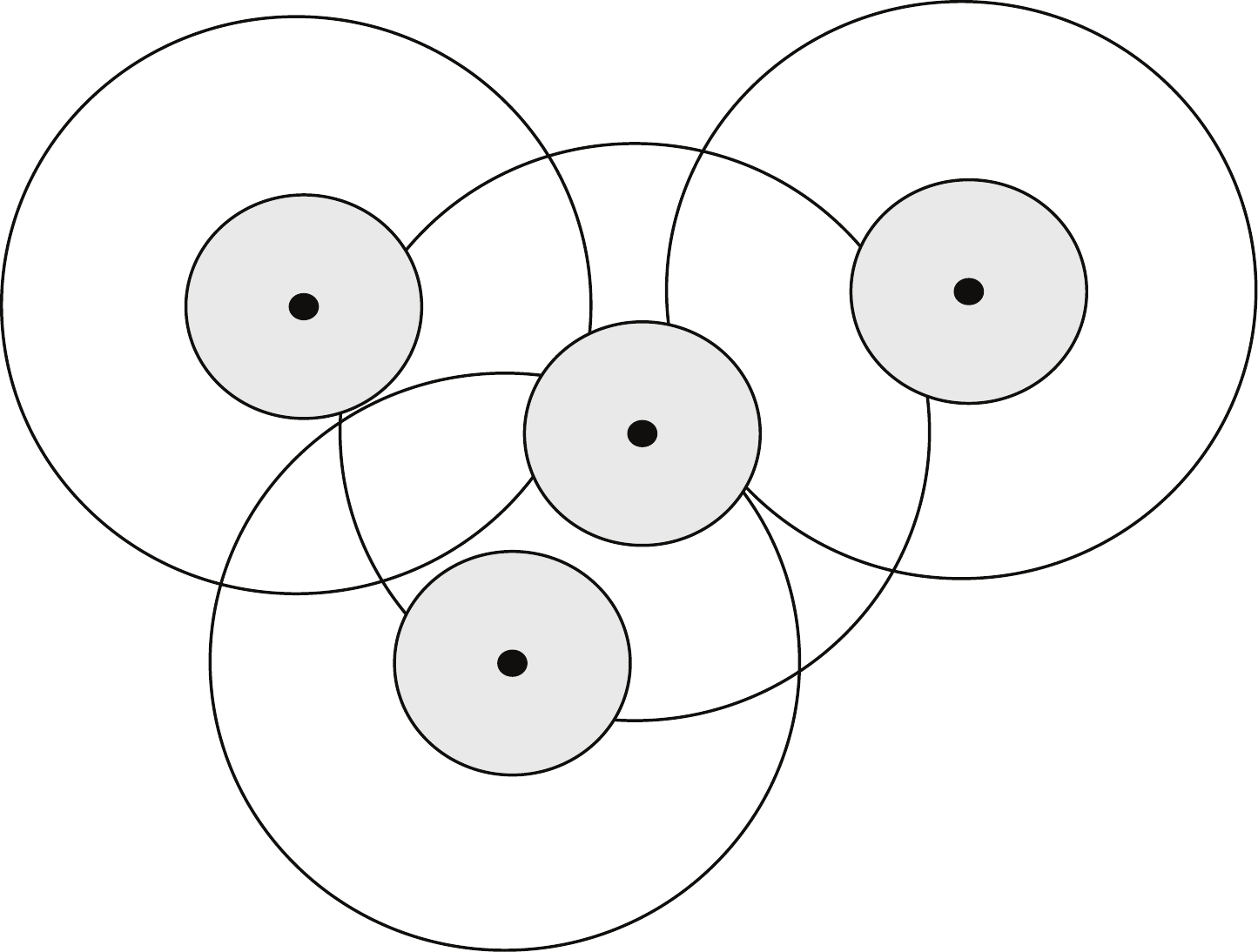}
\caption{An illustration of overlapping interference regions in
neighbouring WLANs. Black dot indicates location of a wireless
access point, shaded area indicates region within which communication
with access point can take place and outer circles indicate regions
within which transmissions interfere.}
\label{fig:example}
\end{figure}

802.11 WLANs are ubiquitous, but their deployments are unstructured
and they operate in an unlicensed frequency band. Within this band,
an 802.11 WLAN can select one of several channels, typically $11$,
to operate on. In the vast majority of WiFi routers channel
selection is, at present, based on operator selection without any
quantitative instruction from the router. Self-selection of this
channel is the task we consider here. Co-ordinated selection is
hampered by the fact that the interference range of a typical 802.11
device is substantially larger than its communication range.
Consequently, WLANs can interfere but may be unable to decode each
others messages (illustrated schematically in Fig. \ref{fig:example}).
Discovery by a WLAN of the existence of interfering WLANs via its
wired back-haul may be prevented by firewalls and, in any case, these
WLANs may not know their physical location. These practical
restrictions mandate a decentralized channel-selection algorithm.

We can identify the task of each WLAN self-selecting a non-interfering
channel with a CSP with communication constraints imposed by the
network topology. Let $x_i$ be the channel selected by WLAN
$i\in\{1,\ldots,N\}$ and define $M=N(N-1)/2$ clauses, one for each
pair of WLANs, that evaluates to one if the WLANs are on non-interfering
channels or are out of interference range and is zero otherwise.
Due to the unstructured nature of the deployment, the lack of shared
administrative control and communication constraints, a WLAN cannot
rely on knowing the number or the identity of interfering WLANs or
the channels that they currently have selected. A practical CSP
solver for this task can only rely on each WLAN being able to measure
whether: (i) all of the neighbouring WLANs have selected a different
channel from it; or (ii) if one or more neighbours have selected
the same channel.

In certain settings, a limited amount of communication of variable
values between the stations hosting them may be feasible. For
example, it may be possible for stations hosting a link or for a
WLAN to overhear beacons or traffic from a subset of its interferers,
or for an access point to communicate with other, nearby access
points. Algorithms for channel selection that are proposed in the
literature assume the existence of end-to-end communication for
centralized solutions, for message passing in decentralized or
gossiping solutions, or to co-ordinate global restart in simulated
annealing proposals e.g. \cite{Wu_06, Kauffmann07, Crichigno08,
Subramanian08} and references therein. This information, however,
tends to be opportunistic in nature and cannot be relied on \emph{a
priori}. We leave how best to exploit such partial information to
future work, focusing here on the most challenging, fundamental
cases where fully decentralized algorithms are required.

\subsection{Contribution}
The primary contributions of this paper are fourfold. Firstly, we show
that CSPs provide a unifying framework that encompasses several
important problems in wireless networks, including channel selection,
inter-session network coding and decentralized scheduling in
a WLAN. Each of these problems is generally thought of as being 
distinct and requiring different solution approaches. We show
that they are fundamentally related as CSPs so that the decentralized
CSP solver introduced in this paper can be used for all of them.
Secondly, we define a new class of CSP algorithms which we term
decentralized CSP solvers and, extending our earlier approach to
graph colouring \cite{leith06}\cite{duffy08}, introduce a
novel stochastic decentralized CSP solver proving that it will find
a solution in almost surely finite time, should one exist, and
showing it has many practically desirable properties. Thirdly, we
benchmark the algorithm's performance on a well-studied class of
application-agnostic CSPs, random k-SAT.  For problems with up to
a thousand variables, which are large for our motivating examples,
we find that the time the algorithm takes to find a satisfying
assignment is close to that of WalkSAT \cite{Selman95}, a well
regarded, efficient centralized CSP solver, while also possessing
desirable features of Survey Propagation \cite{Braunstein05}. That
is, despite its decentralized nature, the algorithm is fast. Fourthly,
we demonstrate the solver's practical utility in a complex, wireless
resource allocation case study.

The rest of the paper is organized as follows. In Section
\ref{subsec:dCSPs} we define decentralized CSP solvers and show
that well-known algorithms for solving CSPs fail to meet these
criteria.  In Section \ref{sec:CFL} we introduce a decentralized
CSP solver.  We prove that it finds a solution in finite time with
probability one whenever a feasible solution exists, obtaining an
upper bound on the algorithm's convergence rate in the process. In
Section \ref{sec:sim} we consider its speed at identifying solutions
of instances of random k-SAT.  In Section \ref{sec:example} we
demonstrate the algorithm's utility in solving the problems that
motivated its introduction. Section \ref{sec:disc} contains discussion
and closing remarks.

\section{Decentralized CSP solvers}
\label{subsec:dCSPs}

We begin by formalizing the criteria that CSP solvers must possess
to be of use for problems with communication constraints such as
those outlined above. We call algorithms that meet these criteria
decentralized CSP solvers.

\begin{definition}[CSP]
A CSP with $N$ variables, $\{x_1,\ldots,x_N\}$, and $M$ clauses is
defined as follows. The variables each take values in a finite set 
$\Dc=\{1,\ldots,D\}$
and $\vx:=(x_1,\ldots,x_N)\in\Dc^N$. Each clause $m\in\Mc=\{1,\ldots,M\}$
is defined by a function $\Phi_m:\Dc^N\mapsto\{0,1\}$ where for an
assignment of variables, $\vx$, $\Phi_m(\vx)=1$ if clause $m$ is
satisfied and $\Phi_m(\vx)=0$ if clause $m$ is not satisfied. An
assignment $\vx$ is a solution to the CSP if all clauses are
simultaneously satisfied. That is,
\begin{align}
\label{eq:CSP}
\vx\text{ is a satisfying assignment iff }
\min_{m\in\Mc} \Phi_m(\vx)=1.
\end{align}
\end{definition}
This encompasses all of our examples of interest as well as k-SAT,
the most well-studied general class of CSPs.   

A CSP solver is an algorithm that can find a satisfying variable
assignment for any solvable CSP.

\begin{definition}[CSP Solver] A CSP solver takes a CSP as input
and determines a sequence $\{\vx(t)\}$ such that for any CSP that
has satisfying assignments:
\begin{description}
\item[\findsoln]
for all $t$ sufficiently large $\vx(t)=\vx$ for some satisfying
assignment $\vx$;
\item[\sticksoln]
if $t'$ is the first entry in the sequence $\{\vx(t)\}$ such that
$\vx(t')$ is a satisfying assignment, then $\vx(t)=\vx(t')$  for
all $t>t'$.
\end{description}
\end{definition}

To provide criteria that classify CSP solvers as being decentralized,
we begin with the following definition.

\begin{definition}[Clause participation]
We say that variable $x_i$ participates in clause $\Phi_m(\vx)$ if
the value of $x_i$ can influence the clause's satisfaction for at
least one assignment of the rest of the variables in $\vx$. 
\end{definition}

For each variable $x_i$, let $\Mc_i$ denote the set of clauses in
which it participates:
\begin{align*}
\Mc_i = 
\bigcup_{x_i\in\{1,\ldots, d\}} \left( 
\bigcup_{j\neq i} \bigcup_{x_j \in \{1,\ldots, d\}}
        \left\{m: \Phi_m(x_i, \{x_j\}) = 0 \right\}\right. \\
\bigcap
\,
\left.
\bigcup_{j\neq i} \bigcup_{x_j \in \{1,\ldots, d\}}
	\left\{m: \Phi_m(x_i, \{x_j\}) = 1 \right\}
	\right).
\end{align*}

Re-writing the left hand side of eq. \eqref{eq:CSP}
in a way that focuses on the satisfaction of each variable we have:
\begin{align}
\label{eq:dCSP}
\vx\text{ is a satisfying assignment iff } \min_i \min_{m\in\Mc_i}
\Phi_m(\vx) = 1.
\end{align}
A decentralized CSP algorithm can be thought of as having 
intelligence co-located with each variable. The intelligence at
variable $x_i$ can only determine whether all of the clauses that
$x_i$ participates in are satisfied or that at least one clause is
unsatisfied and must make update decisions locally based solely on this
knowledge. 
\begin{definition}[Decentralized CSP Solver]
A decentralized CSP solver is a CSP solver that for each variable
$x_i$, $i\in\{1,\ldots,N\}$, must select its next value based only
on an evaluation of 
\begin{align}
\label{eq:minPhi}
\min_{m\in\Mc_i} \Phi_m(\vx).
\end{align}
That is, for each variable $x_i$ the solver must make a decision
on the next value of $x_i$ based solely on knowing whether that
variable is currently satisfied or not without explicitly knowing:
\begin{description} 
\item[\nox]
the assignments of other variables, $x_j$ for any $j\neq i$;
\item[\noM] 
the set of clauses that any variable, including itself, participates
in, $\Mc_j$ for $j\in\{1,\ldots,N\}$;
\item[\noPhi]
the functions that define clauses, $\Phi_m$ for $m\in\{1,\ldots,M\}$.
\end{description} 
\end{definition}
Note that the properties \findsoln \sticksoln of any CSP solver
mean that a decentralized CSP solver must, without explicit
communication, settle upon a satisfying assignment the first time
one is found. Communication to co-ordinate global stopping or
restarting of the solver would be contrary to the nature of the
natural constraints of these problems and so is forbidden.

The stochastic algorithm we define in Section \ref{sec:CFL} shall
provably satisfy the properties of a decentralized CSP solver with
probability one. Moreover, it will also have the desirable property
that it will automatically restart its search, without communication,
upon any change to the clauses of a CSP that makes the current
variable assignment no longer satisfying. This is significant in
practical applications as, for example, the arrival of new transmitters
in a wireless network would induce a change the associated CSP.

\subsection{Formulating Wireless Network Tasks As CSPs}

Before proceeding, we demonstrate that several important resource
allocation tasks in wireless networks fall within this CSP framework,
which, therefore, provides a unifying framework for analyzing these
tasks.

{\it 1) Graph colouring.}  
As briefly described in Section \ref{sec:channel_all}, the channel
assignment problem corresponds to decentralized graph colouring.
In the simplest model, the network is described by an undirected
interference graph $G=(V,E)$ with vertices $V$ and edges $E$. Each
vertex represents a WLAN and an edge exists between two WLANs if
they interfere with each other when on the same channel.
Define $N=|V|$ and $M=|E|$ and let $\Dc$ denote
the set of available colours (channels) and let $x_i$ be a variable
with value equal to the colour selected by an intelligence co-located
with each vertex $i\in V$. Each clause $m\in\{1,\ldots,|E|\}$, which
is an enumeration of the elements of $E$, can be identified with
an edge $(i,j)\in E$. We define
\begin{align*}
\Phi_m(\vx)= \Phi_m(x_i,x_j)=
	\begin{cases} 
	1 & \text{if } x_i\ne x_j\\
	0 & \text{otherwise}.  
	\end{cases}
\end{align*}
The participation set, $\Mc_i$, of a variable $x_i$ consists of
all clauses where the vertex $i$ is one end of the associated edge,
\begin{align*}
\Mc_i = \left\{m\equiv (i,j): (i,j)\in E\right\}.
\end{align*}
A variable assignment satisfies this CSP if and only if it is
also a proper colouring for the graph $G$. Proper
colourings correspond to channel assignments in which no two
neighbouring WLANs in the interference graph have selected the same
channel. To find a proper colouring, when one exists, a decentralized CSP
solver requires only that an intelligence co-located with each
vertex $i$ can measure whether: (i) all of its neighbours are using
a different colour from vertex $i$; or (ii) at least one neighbour
has selected the same colour as vertex $i$. This information is
sufficient to evaluate eq.  \eqref{eq:minPhi} and, in particular,
the intelligence at each vertex does not need to know: \nox the
colour selected by any other vertex; \noM its set of neighbours;
or \noPhi the exact nature of the clauses $\Phi_m$, $m\in\{1,\ldots,M\}$.

{\it 2) Channel assignment with channel-dependent interference.}
Whether or not transmitters interfere in the channel assignment task 
may depend on the radio channel selected. This can arise due to
frequency dependent radio propagation or to channel dependent
spectral masks. That is, for regulatory reasons, different spectral
masks are typically used when transmitting on channels at the edge
and on channels in the middle of a radio band.  This problem then
has a collection of conflict graphs, one for each possible radio
channel. This version of the channel assignment task can also be
formulated as a CSP even though it is no longer a graph colouring
problem. Let $G(c)=(V,E(c))$, $c\in\Dc$ be a set of undirected
graphs with the same vertices $V$ but possibly differing edge sets
$E(c)$. Again each vertex represents a WLAN, but the interference
graph is channel-dependent.
The Graph $G(c)$ is associated with radio channel $c$ and
an edge exists in $E(c)$ if two WLANs interfere while on channel
$c$. Let $x_i$ be a variable with value equal to the channel selected
by each vertex $i\in V$. Each clause
$m\in\{1,\ldots,|E(1)|+\cdots+|E(D)|\}$, which is an enumeration of
all edges in all graphs, can be identified with a $c\in\Dc$ and an
edge $(i,j)\in E(c)$ and is defined by
\begin{align*}
\Phi_m(\vx)= \Phi_m(x_i, x_j)=
	\begin{cases}
	0 & \text{if } x_i = x_j = c\\
	1 & \text{ otherwise.}
	\end{cases}
\end{align*}
The collection of clauses that variable $x_i$, associated with
vertex $i$, participates in can be identified as
\begin{align*} 
\Mc_i = \bigcup_{c\in\Dc}\left\{m\equiv (i,j,c): (i,j)\in E(c)\right\}. 
\end{align*} 
With $x_i$ being the current channel selection of vertex $i$, for
all $c\neq x_i$ the clauses associated with $m\equiv (i,j,c)$ are
automatically satisfied. Thus to evaluate eq. \eqref{eq:minPhi}
it is sufficient for the station to measure if no neighbour coincides
with its current channel selection or if at least one does.
To find an interference-free channel assignment, a decentralized
CSP solver requires that each vertex $i$ can measure whether: (i)
all of its neighbours on the currently selected channel are using
a different channel from itself; or (ii) at least one neighbour has
selected the same channel. 

{\it 3) Inter-session network coding.}
Network coding has been the subject of considerable interest in
recent years as it offers the potential for significant increases
in network capacity \cite{Koetter03, Deb05, Katti06}. In network
coding, network elements combine packets together before transmission
rather than forwarding them unmodified. The combined packets can
be from individual flows, known as intra-flow coding, or across
multiple flows, inter-flow coding.

While intra-flow coding within multicast flows has been well studied,
inter-flow coding between unicast flows has received less attention,
yet is, perhaps, of more immediate relevance to current Internet
traffic. Inter-flow coding is known to be challenging \cite{Sengupta10}.
The task of a network finding a feasible linear network code
in a distributed fashion, but with some global sharing of calculations,
is investigated in \cite{Kim09} through the use of a genetic
algorithm. This task can be formulated as a CSP.

Let $G=(V,E)$ denote a directed acyclic multi-graph representing
the network with vertices $V$ and edges $E$. Each vertex
represents either a source of an information flow, a destination
or router in the network. Edges represent physical connections
between these elements. Time is slotted and each edge can 
transmit a single packet per slot. We allow multiple edges
between vertex pairs in order to accommodate higher rate links.  It
is the goal of each link beyond the source to determine a linear
combination of its incoming packets to forward so that ultimately
all flows get to their destinations. As the combination of packets
coded by each link will take care of routing, each flow $p\in P$
is defined by its source, $\sigma(p)\in V$, and its destination
$\delta(p)\in V$. Each flow is assumed to be unit rate with higher
rate flows accommodated by flows with the same source and destination
vertices. We will treat the collection of source, $\bigcup_{p\in
P}\{\sigma(p)\}$, and destination vertices, $\bigcup_{p\in
P}\{\delta(p)\}$, as special, solely having one incoming and one
outgoing edge per flow respectively.

For each edge $i\in E$, define $s(i)$, $s:E\mapsto V$, to be its
source vertex and $t(i)$, $t:E\mapsto V$, to be its target vertex.
For each vertex $v\in V$ we define the set of incoming edges
$I_v:=\{i\in E: t(i)=v\}$ and the set of outgoing edges $O_v:=\{i\in
E: s(i)=v\}$. For each edge not corresponding to source, $i\in
E\setminus \bigcup_{p\in P}O_{\sigma(p)}$, we associate a variable
$x_i\in\Dc:=\{1,\ldots,2^{|P|}\}$, which is an enumeration of
the power set of the set of flows $P$. We define a bijective map 
$\psi:\Dc\mapsto\{0,1\}^{|P|}$ that determines a vector whose
positive entries correspond to flows to be coded. 

In total, we have $|E\setminus\bigcup_{p} O_{\sigma(p)}|$ clauses.
One for every link that isn't an outgoing link from a source vertex.
At each time slot, each link that doesn't correspond to a source or
final destination link,
$i\in E\setminus\bigcup_{p}(O_{\sigma(p)}\cup I_{\delta(p)})$,
wishes to forward a packet consisting of XORed packets (corresponding
to addition in the Galois field of two elements) from the flows
indicated by $\psi(x_i)$. Each final link $i\in\bigcup_{p}
I_{\delta(p)}$ wishes to forward packets from $p$, which we
indicate by the vector $\gamma_p$ with a $1$ at the location $p$
and zeros elsewhere. A link and its immediate upstream neighbouring
links will be dissatisfied if it cannot do so.

More formally, for each edge not corresponding to
the outgoing link from a source or incoming link to a destination,
$i\in E\setminus \bigcup_{p\in P}(O_{\sigma(p)}\cup I_{\delta(p)})$,
we have a clause $m\equiv i$, $\Phi_m(\vx)$, defined by
\begin{align*}
\Phi_m(x_i,x_{j:j\in I_{s(i)}})
	=
	\begin{cases}
	1 & \text{if } \exists\, \theta \text{ s.t. }
		\Psi(I_{s(i)}) \theta^T= \psi(x_i)^T\\	
	0 & \text{otherwise}.
	\end{cases}
\end{align*}
where $\theta$ is a binary vector and $\Psi(I_{s(i)})$ is the
rectangular matrix consisting of $\psi(x_j)^T$ for $j\in
I_{s(i)}\setminus\bigcup_{p\in P}O_{\sigma(p)}$ and
and $\gamma_p^T$ if $i\in O_{\sigma(p)}$ for some $p$. 

To complete the CSP, we need to ensure that packets from flow
$p$ can be decoded at their destination $\delta(p)$. For 
$i\in\bigcup_{p\in P}I_{\delta(p)}$, then with $\delta(p')=t(i)$
we introduce a clause 
\begin{align*}
\Phi_m(x_{j:j\in I_{s(i)}})
	=
	\begin{cases}
	1 & \text{if } \exists\, \theta \text{ s.t. }
		\Psi(I_{s(i)})\theta^T = \gamma_{p'}^T\\	
	0 & \text{otherwise}.
	\end{cases}
\end{align*}
where again $\theta$ is a binary vector and $\Psi$ is is the
rectangular matrix defined above. The set of clauses a link variable
$x_i$, $i\in
E\setminus (\bigcup_{p\in P}O_{\sigma(p)} \cup I_{\delta(p)})$,
participates in is
\begin{align*}
\Mc_i =\{ m \equiv j: j=i \text{ or } i\in I_{s(j)}\}
\end{align*}
Any variable assignment then satisfies this CSP if and only if it
is a realizable network code satisfying all flow demands.

Hence, to find a proper assignment a decentralized CSP
solver requires only that each link $i$ can determine whether: (i)
its own coding is realizable and the coding of its immediately
down-stream links are realizable; or (ii) if at least one of these
is not realizable. This is sufficient to evaluate eq. \eqref{eq:minPhi}.
Each link does not need to explicitly know: \nox the code selected
by any other link; \noM the network topology; \noPhi any details
of how code realizability is determined at any link.

{\it 4) Decentralized transmission scheduling.} In an Ethernet or
WLAN it is necessary to schedule transmissions by stations. This
might be achieved in a centralized manner using TDMA, but it can
also be formulated as a decentralized problem.  The classical CSMA/CA
approach to decentralized scheduling never settles to a single
schedule and some comes at a cost of the possibility of continual
collisions.  Recently, there has been interest in decentralized
approaches for finding collision-free schedules, see
\cite{barcelo10,fang12}. This task can be formulated as a CSP as
follows. Let $V$ denote the set of transmitters in the WLAN, $T$
denote the set of available time slots and define $N=|V|$ and
$M=N(N-1)/2$. Let $x_i$ be a variable with value equal to the
transmission slot selected by transmitter $i\in V$. Define a clause
$\Phi_m(\vx)$ associated with each pair of transmitters $m\equiv
(i,j)$ such that
\begin{align*}
\Phi_m(x_i,x_j)=
	\begin{cases}
	1 & \text{ if } x_i\ne x_j  \\
	0 & \text{ otherwise}. 
	\end{cases}
\end{align*}
The participation sets are 
\begin{align*}
\Mc_i=\bigcup_{j\neq i}\{m \equiv (i,j)\}.
\end{align*}
Any variable assignment satisfies
this CSP if and only if it is also a collision-free time-slot
schedule. To find a collision-free schedule, when one exists, a
decentralized CSP solver requires only that each transmitter $i$
can measure whether: (i) all of its neighbours are using a different
time-slot from transmitter $i$; or (ii) at least one transmitter
in the WLAN has selected the same time-slot as transmitter $i$.
Again, this is all that is needed to evaluate eq. \eqref{eq:minPhi}.
Each transmitter does not need to know: \nox the time-slot selected
by any other transmitter; \noM the set of transmitters; \noPhi the
clauses.

\subsection{Related work - existing algorithms are not decentralized}

The literature on general purpose CSP solvers is vast, typically
focusing on solving k-SAT problems in conjunctive normal form, but
they can be broadly classified into those based on: (i) the
Davis-Putnam-Logemann-Loveland (DPLL) algorithm \cite{Davis60,
Davis62}; (ii) Survey Propagation \cite{Braunstein05}; and (iii)
on Stochastic Local Search (SLS) \cite{Selman95}. Each of these
approaches has experienced substantial development and has its own
merits, but none were motivated by problems where variables
have a geographical sense of locality.

Algorithms developed from the DPLL approach have proved to be the
quickest at SAT-Race and SAT Competition in recent years, e.g.
ManySAT \cite{Hamadi09}. The DPLL approach ultimately guarantees a
complete search of the solution space and so meets the \findsoln
and \sticksoln criteria. They are, however, based on a branching
rule methodology, e.g. \cite{Ouyang98}, that assumes the existence
of a centralized intelligence that employs a backtracking search.
The implicit assumptions of the information available to this
intelligence breaks the conditions \nox \noM \noPhi and so they are
not decentralized CSP solvers.

Survey propagation, a development of belief propagation \cite{pearl82}
from trees to general graphs, has proved effective in graphs that
do not contain small loops \cite{mezard02a}. For a given CSP, the
fundamental structure of study is a called a factor graph. In order
to generate this, it is necessary to know what clauses each variable
participates in and the nature of each of the clauses, breaking the
\noM and \noPhi criteria and so these are not decentralized CSP solvers.

SLS algorithms also depend fundamentally upon the exchange of
information, mostly in an explicit manner breaking the \nox condition
by basing update decisions on relative rankings of the constraint
variables but also in a more subtle fashion. To see this implicit
requirement, consider the following algorithm for binary valued
variables originally proposed by Papadimitriou \cite{papadimitriou91}
and developed further by Sch\"oning \cite{schoning99}. Pick a random
assignment of values for the constraint variables.  Repeat the
following: from all of the unsatisfied clauses, pick one uniformly at
random, select one of the variables participating in that clause 
and negate its value, breaking the \noM and \noPhi conditions.
The algorithm halts when all clauses are satisfied or a specified time
limit expires.  Although simple, this forms the basic building block
for all SLS algorithms, including the well-studied WalkSAT algorithm
\cite{Selman95}. It is important that a single unsatisfied clause
is selected at each step and that a single variable within the
clause is adjusted as it is this that leads to the algorithm behaving
as a random walk \cite{schoning99}. Thus, again, solvers in this
class are not decentralized.

\section{A decentralized CSP solver}
\label{sec:CFL}

We now introduce an algorithm that satisfies the decentralized CSP
solver criteria.

\subsection{Communication-Free Learning Algorithm}\  
An instance of the following Communication-Free Learning (CFL)
algorithm is run in parallel for every variable. For each variable,
$i\in\{1,2,...,N\}$ it keeps a probability distribution,
$p_{i,j}$ over $j\in D$ as well as the current variable value $x_i$.
In pseudocode, the CFL algorithm is:
\begin{algorithm}
\caption{Communication-Free Learning}
\label{CFL}
\begin{algorithmic}[1]
\STATE Initialize $p_{i,j}=1/d$, $j\in\{1,...,D\}$.
\LOOP
\STATE 
Realize a random variable, selecting $x_i=j$ with probability $p_{i,j}$.  
\STATE 
Evaluate $\min_{m\in\Mc_i} \Phi_m(\vx)$, returning \emph{satisfied}
if its value is $1$, as this indicates all of variable $i$'s clauses
are satisfied given the present assignment, and \emph{unsatisfied}
otherwise.
\STATE  
Update:  
If \emph{satisfied},
\begin{align*}
p_{i,j} & = 
\left\{\begin{array}{cc}
1 & \text{if $j=x_i$} \\
0 & \text{otherwise}.
\end{array}\right.
\end{align*}
If \emph{unsatisfied},
\begin{align*}
p_{i,j} =
\begin{cases}
(1-b)p_{i,j} +a/(D-1+a/b) &\text{ if }  j= x_i\\
(1-b)p_{i,j} +b/(D-1+a/b) &\text{ otherwise},
\end{cases}
\end{align*}
where $a,b\in(0,1]$ are design parameters.
\ENDLOOP
\end{algorithmic}
\end{algorithm}

To understand the logic behind CFL, note that for each variable
a probability distribution over all possible variable-values is
kept. The variable's value is then selected from this distribution.
Should all the clauses that the variable participates in be satisfied
with its current value, the associated probability distribution is
updated to ensure that the variable value remains unchanged. If at
least one clause is unsatisfied, then the probability distribution
evolves by interpolating between it and a distribution that is
uniform on all values apart from on the one that is presently
generating dissatisfaction. Thus if all of a variable's clauses
were once satisfied, the algorithm retains memory of this into the
future. This has the effect that if a collection of variables are
content, they can be resistant, but not impervious, to propagation
of dissatisfaction from other variables.

CFL possesses two parameters. The parameter $b$ determines how
quickly the past is forgotten, while $a$ determines the algorithm's
aversion to a variable value should it be found to cause dissatisfaction.
Even though an instance of CFL is run for each variable, we shall show
that this completely decentralized solution is a CSP solver.

From here on we will assume that the update Step 5 is performed in
a synchronized fashion across variables. Without synchronization,
the fundamental character of the algorithm doesn't change, but the
analysis becomes more involved. Synchronization solely requires
that algorithm instances each have access to a shared sense of time
and that this can be achieved without information-sharing or other
communication between variables, i.e.  between algorithm instances.
A suitable clock is, for example, available to any Internet connected
device via the Network Time Protocol (NTP).

\subsection{CFL is a decentralized CSP solver}

By construction, the only information used by the algorithm is
$\min_{m\in\Mc_i}\Phi_i(\vx)$ in Step 4 and thus it satisfies the
criteria \nox, \noM and \noPhi. That is, it only needs to know if
all clauses in which variable $i$ participates are satisfied or if
one or more are not. The CFL algorithm also satisfies the \sticksoln
criterion that it sticks with a solution from the first time one
is found. To see this, note that the affect of Step 5 is that if a
variable experiences success in all clauses that it participates
in it continues to select the same value with probability $1$. Thus
if all variables are simultaneously satisfied in all clauses, i.e.
if $\min_i\min_{m\in\Mc_i}\Phi_i(\vx)=1$, then the same assignment
will be reselected indefinitely with probability $1$.

To establish that the CFL algorithm is a decentralized CSP solver
all that remains is to show that it meets the \findsoln criterion,
that if the problem has a solution it will be found, which is dealt
with by the the following theorem. It provides an upper bound on
the distribution of the number of iterations the algorithm requires
to find a solution to any solvable CSP. It's proof exploits the
iterated function system structure of the algorithm and can be found
in the Appendix.

\begin{theorem}
\label{thm:main}
For any satisfiable CSP, with probability greater than $1-\epsilon\in(0,1)$
the number of iterations for the CFL algorithm to find a satisfying
assignment is less than
\begin{align*}
N\exp\left(\frac{N(N+1)}{2}\log(\gamma^{-1})\right)\log(\epsilon^{-1}),\\
\text{ where }
\gamma = \frac{\min(a,b)}{D-1+a/b}.
\end{align*}
For a CSP corresponding to graph coloring, a tighter bound holds
and a satisfying assignment will be found with probability greater
than $1-\epsilon$ in a number of iterations less than
\begin{align*}
N\exp(2N\log(\gamma^{-1}))\log(\epsilon^{-1}).
\end{align*}
\end{theorem}

Theorem \ref{thm:main} proves that for fixed $N$ the tail of the
distribution of the number of iterations until the first identification
of a satisfying assignment is bounded above by a geometric distribution
and so all of its moments are finite. As Theorem \ref{thm:main}
covers any arbitrary CSP that admits a solution, for any given
instance these bounds are likely to be loose. They do, however,
allow us to conclude the following corollary proving that if a
solution exists, the CFL algorithm will almost surely find it.
\begin{corollary} For any CSP that admits a satisfying assignment,
the CFL algorithm will find a satisfying assignment in almost surely
finite time.
\end{corollary}

Hence the CFL algorithm satisfies all of the criteria \findsoln
\sticksoln \nox \noM \noPhi, almost surely, and so is a decentralized
CSP solver.

\subsection{Parameterization}
Theorem \ref{thm:main} establishes that the CFL algorithm provably
identifies satisfying assignments for all values of its two design
parameters, $a$ and $b$. The value of $a$ determines the algorithm's
aversion to variable values for which clause failure has been
experienced. The value of $b$ impacts on the speed of convergence
of the algorithm. Optimal values of $a$ and $b$ depend upon
each problem and a performance metric. For fast convergence across
a broad range of CSPs with distinct structure, we have found that
small values of $a$ and $b$, corresponding to strong aversion to
a dissatisfying variable value and reasonably long memory, 
work well. Thus for simplicity, we set $a=b$ in all experimental
examples. We use $b=0.2$ for random 3-SAT, $b=0.1$ for random 4-SAT
and $b=0.05$ for random 5-SAT. We use $b=0.1$ for our wireless
networks example, a value which we have found to yield good performance
across a range of $k$-SAT problems.

\section{Application-agnostic benchmarking}
\label{sec:sim}

\begin{table*}
\begin{center}
\begin{tabular}{|l|c|c|c|c|c|c|c|} 
\hline
$k$                          & 3         &  4       &5 &        7 &    10   &   20      \\
\hline
Theoretically derived bounds		     &		&	&	&	&	& \\
\hline
$2^k \log2$       	& 5.54   & 11.09 &22.18  &88.72 &709.78 & 726,817   \\
$r_{\neg exist,k}$\cite{Achlioptas05} & 4.51   & 10.23 &21.33  &87.88 &708.94 &726,817  \\
$r_{exist,k}$\cite{Achlioptas04,Kaporis06}           & 3.52   & 7.91   &18.79  &84.82 &704.94 &726,809  \\
$2^k \log2-k$         & 2.54   & 7.09 &17.18  &81.72 &699.78 & 706,817   \\
$r_{poly,k}$ \cite{Frieze96,Kaporis06}          & 3.52   & 5.54  &9.63     &33.23 &172.65& 95,263  \\
\hline
Estimates		     &		&	&	&	&	& \\
\hline
$r_{\neg exist,k}$\cite{Mertens05}  &4.267   & 9.93    & 21.12 & 87.79 &     708.91      &      -       \\
$r_{1RSB,k}$\cite{Mertens05,Krzakala07}    & 4.15   & 9.38   &19.16    &62.5 &         -        &       -      \\
\hline
\end{tabular} 
\caption{Current best upper/lower bounds on the phase transition
thresholds in random k-SAT.} 
\label{table}
\end{center}
\end{table*}

A CSP associated with a specific networking task will possess
structure induced by its topology. Before considering practically-motivated
applications in Section \ref{sec:example}, as the CFL algorithm can
solve any CSP but its design was subject to restrictions not normally
considered it is prudent to investigate its performance as a CSP
solver in an application-agnostic setting.

We present simulation data evaluating the performance of the CFL
algorithm for random k-SAT, a well-studied class of CSPs.  A k-SAT
problem is a CSP in which all variables are binary-valued and clauses
consist solely of logical disjunctions of no more than k variables.
In random k-SAT, an instance of k-SAT is generated by drawing $M$
such clauses uniformly at random \cite{papadimitriou91}.  Our
investigation surprisingly reveals that the CFL algorithm is
competitive with centralized and distributed solvers on problems
of reasonable size (order 1000 variables).

Briefly, we first review current knowledge regarding
random k-SAT. The behavior of random k-SAT is known to depend
strongly on the parameter $r=M/N$ with  phase
transitions and associated thresholds $r_{\neg{exist},k}$ and
$r_{{exist},k}$. If $r>r_{\neg{exist},k}$ the constraint problem
is unsatisfiable with high probability, while if $r<r_{exist,k}$
a satisfying assignments exist with high probability. Evidently,
$r_{{exist},k}\le r_{\neg{exist},k}$ and it is conjectured that
$r_{{exist},k} =r_{\neg{exist},k}$ \cite{Achlioptas04}. A simple
argument gives $r_{\neg{exist},k}\le 2^k \log 2$ and this upper
bound can be refined to obtain the values in the second row of Table
\ref{table}.
Also shown in row four of this table are estimated values for
$r_{\neg{exist},k}$ derived from statistical physics considerations.
These latter estimates are supported by experimental data
for $k=3$, e.g. \cite{NIPS04}, although there are fewer
experimental studies for $k>3$. It can be seen that the theoretical
bound for $r_{\neg{exist},k}$ and the estimated values are in good
agreement for $k\ge 5$, and that both approach $2^k \log 2$ for
large $k$. Recent mathematical results have established that
$r_{exist,k} \ge 2^k \log 2-k$ \cite{Achlioptas05} and this lower
bound can be tightened to obtain the theoretical values shown in
the third row of Table \ref{table}.

There also exists a threshold $r_{poly,k}$ below which a satisfying
assignment can be found in polynomial time with high probability.
This threshold has been the subject of much interest and the current
best analytic lower bound for $r_{poly,k}$ is also indicated in
Table \ref{table}. Statistical physics considerations have led to
the conjecture that $r_{poly,k}$ is equal to the value $r_{1RSB,k}$
at which the one-step replica symmetry breaking (1RSB) instability
occurs \cite{Mezard02}.  Current best estimates for this value are
given in Table \ref{table}.  For $k\ge 8$, it has been proven
analytically that the set of satisfying assignments is grouped into
widely separated clusters, which lends support to this conjecture
\cite{Mezard05}. For values of $k<8$ the situation is less clear,
with experimental evidence indicating that $r_{poly,k}$ lies above
$r_{1RSB,k}$ for $k=3,4$ and $5$ \cite{Alava08}.

\begin{figure}
\centering
\includegraphics[width=3in]{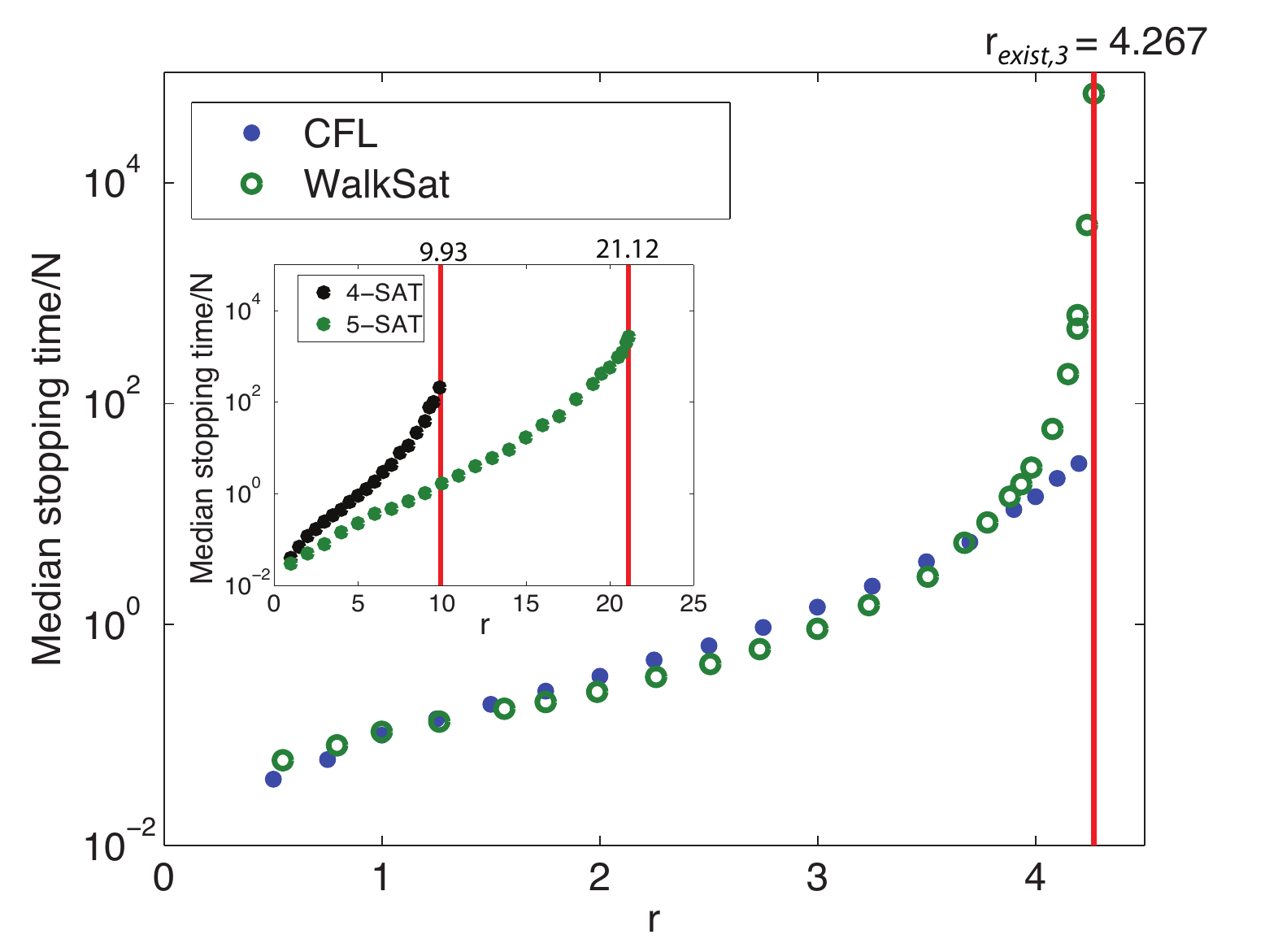}
\caption{Median normalized stopping time of CFL algorithm vs $r$
for random 3-SAT and $N=100$. Each point is derived from at
least 1000 runs of the algorithm; algorithm terminated after $10^7$
iterations if no satisfying assignment found. For comparison,
stopping time measurements for WalkSAT taken from \cite[Fig 2a]{NIPS04}
are also indicated. (\emph{Inset}) CFL algorithm measurements for
random 4-SAT and 5-SAT. Measurements shown are for values of $r$
up to $4.20$ (3-SAT), $9.90$ (4-SAT), $21.10$ (5-SAT); close to the
current best estimates for $r_{{exist},k}$ and well above $r_{1RSB,k}$.}
\label{fig:ksatvsR}
\end{figure}

\begin{figure}
\centering
\includegraphics[width=3in]{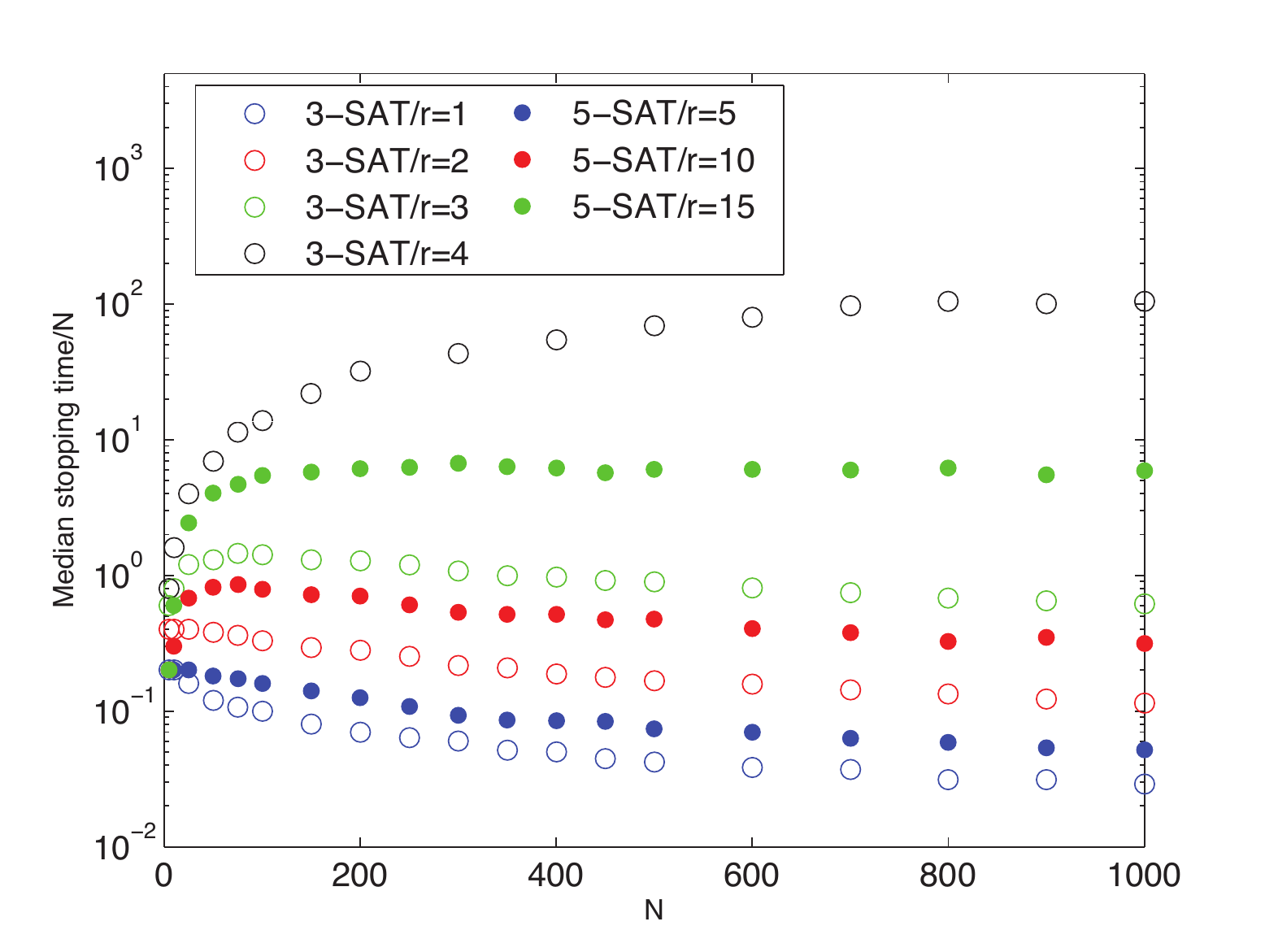}
\caption{Normalized stopping time of CFL algorithm vs $N$ and $r$
for 3-SAT and 5-SAT. Each point is derived from at least 1000
runs.}\label{fig:ksatvsN}
\end{figure}

We now consider the performance of the CFL algorithm. Fig.
\ref{fig:ksatvsR} gives median measurements of normalized 
stopping time (stopping time divided by $N$) of the CFL algorithm
vs $r$=M/N when the number of variables $N=100$. Fig.  \ref{fig:ksatvsN}
gives median measurements of normalized stopping time vs $N$ with
$r$ held constant. Data is shown for random k-SAT with $k=3, 4$ and
$5$. Observe two striking features from this data.  Firstly, from
Fig.  \ref{fig:ksatvsR} noting that a log scale is used on the
$y$-axis, the median normalized stopping time increases exponentially
with $r$ when $N$ is held constant .  Secondly, from Fig.
\ref{fig:ksatvsN}  the median normalized stopping time vs $N$, with
$r$ held constant, is upper bounded by a constant.  That is, a
satisfying assignment is found in a time with median value that
increases no more than linearly with $N$. This linearity holds even
when $r$ is close to $r_{{exist},k}$ (data is shown in Fig.
\ref{fig:ksatvsN} for 3-SAT with $r=4$) and is of great practical
importance as it implies that with high probability the CFL algorithm
finds a satisfying assignment in polynomial time.

Comparing the performance of the CFL algorithm with the popular
WalkSAT algorithm, we note that the WalkSAT algorithm also exhibits
an exponential-like dependence of stopping time on $r$ and linearity
in $N$ \cite{NIPS04,Alekhnovich06,Alava08}. To allow comparison
in more detail, median stopping time data taken from \cite[Fig
2a]{NIPS04} is marked on Fig. \ref{fig:ksatvsR}. For $r<3.9$, 
the median stopping time is similar for CFL and WalkSAT.
Above this value, however, the stopping time for WalkSAT diverges,
increasing super-exponentially in $r$. This divergence is a
feature not only of WalkSAT but also of other local search algorithms
algorithms \emph{e.g.} ChainSat \cite{Alava08}. It is not exhibited
by the Survey Propagation algorithm \cite{NIPS04,Braunstein05},
which has been the subject of considerable interest as 
it creates the ability to operate close to the $r_{{exist},k}$
threshold. Observe that the CFL algorithm also does \emph{not}
exhibit divergence as $r$ approaches $r_{{exist},k}$. These
comparisons are encouraging as they indicate that the CFL algorithm
is competitive with some of the most efficient general-purpose
k-SAT algorithms currently available. It is also unexpected
as information sharing is a key component of both WalkSAT and
Survey Propagation, whereas CFL makes local decisions simultaneously
with no use whatsoever of information-sharing, raising fundamental
questions as to the role of message passing, the relationship between
information exchange and algorithm performance and, in particular,
what performance cost is imposed by constraining attention to
decentralized operation.


\section{Case study: channel allocation in 802.11 WLANs}
\label{sec:example}
%
%

\begin{figure}
\centering
\includegraphics[width=0.8\columnwidth]{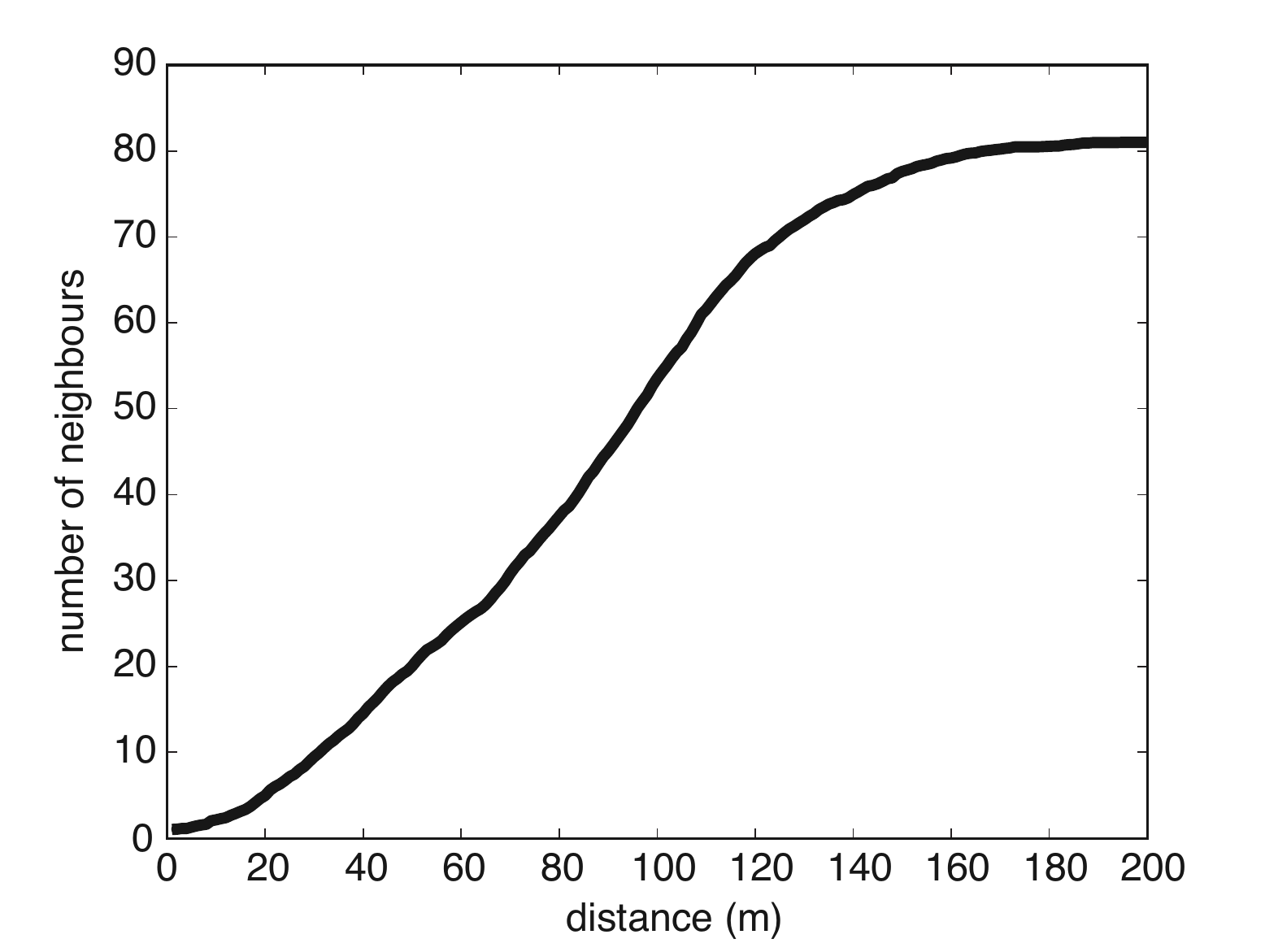}
\caption{Mean number of APs within distance $d$ of an AP vs distance $d$.}\label{fig:meandistance}
\end{figure}

Having established general properties of CFL and determined its
performance on random instances of k-SAT, we return to a problem
of the sort that motivated its introduction. We consider the
performance of the CFL algorithm in a realistic wireless network
case study. From the online database WIGLE \cite{wigle} we obtained
the locations of WiFi wireless Access Points (APs) in an approximately
$150m^2$ area at the junction of 5th Avenue and 59th Street in
Manhattan\footnote{The extracted (x,y,z) coordinate data used is
available online at \url{www.hamilton.ie/net/xyz.txt}}. This
space contains $81$ APs utilizing the IEEE 802.11 wireless
standard.  It can be seen from Figure \ref{fig:meandistance}, which
plots the mean number of APs lying within distance $d$ of an AP,
that within a 15m radius an AP has on average 3 neighbours and
within a 30m radius it has on average 10 neighbours. Of the 11
channels available in the 802.11 protocol, only 3 are orthogonal.
Thus managing interference in this dense deployment is a challenging
task.

Imagine a worst-case scenario where after a power-outage all these
APs are switched back on. The aim of each AP is to select its radio
channel in such a way as to ensure that it is sufficiently different
from nearby WLANs. This can be written as a CSP where we have
$N=81$ APs and $N$ variables $x_i$ corresponding to the channel of
AP $i$, $i=1,2,...,N$.  As per the 802.11 standard \cite{802.11}
and FCC regulations, each AP can select from one of 11 radio channels
in the 2.4GHz band and so the $x_i$, $i=1,2,..,N$ take values in
$\Dc=\{1,2,..,11\}$.  To avoid excessive interference each AP
requires that: (a) no other AP within a 5m distance operates closer
than 3 channels away; (b) no AP within a 10m distance operates
closer than 2 channels away and (c) no AP with a 30m distance
operates on the same channel. We can realize this as a CSP with
$3$ constraint clauses per AP, giving $3N$ in total,
$\Phi_m(\vx)$, $m=1,...,3N$. With $e(i,j)$ denoting the Euclidean
distance between APs $i$ and $j$ in metres, for $m=1,...,N$
\begin{align*}
\Phi_m(\vx)= 
	\begin{cases}
	1 & \text{if } \min_{j:e(m,j)<5}|x_m-x_j|\ge 3 \\
	0 & \text{otherwise},
	\end{cases}
\end{align*}
for $m=N+1,...,2N$
\begin{align*}
\Phi_m(\vx)= 
	\begin{cases}
	1 & \text{if } \min_{j:e(m-N,j)<10}|x_{m-N}-x_j|\ge 2 \\
	0 & \text{otherwise},
	\end{cases}
\end{align*}
and for $m=2N+1,...,3N$
\begin{align*}
\Phi_m(\vx)= 
	\begin{cases}
	1 & \text{if } \min_{j:e(m-2N,j)<30}|x_{m-2N}-x_j|\ge 1 \\
	0 & \text{otherwise},
	\end{cases}
\end{align*}
The attenuation between adjacent channels is $-28$dB \cite{802.11}.
Taking the radio path loss with distance as  $d^\alpha$, where $d$
is distance in meters and $\alpha=4$ the path loss exponent, then
these constraints ensure $>60$ dB attenuation between APs.

Assuming all APs use the maximum transmit power of 18dBm allowed by the 802.11 standard, this means that the SINR is greater than 20dB within a 10m radius of each AP which is sufficient to sustain a data rate of 54Mbps when the connection is line of sight and channel
noise is Gaussian \cite{802.11,goldsmith05}.

Observe that each variable $x_i$ is located at a different AP. The
APs do not belong to a single administrative domain and so security
measures such as firewalls prevent communication over the wired
network. An AP cannot rely on decoding wireless transmissions to
communicate with all of its interferers or even just to identify
them. This is because interferers may be too far away to allow their
transmissions to be decoded and yet still their aggregate transmission
power may be sufficiently powerful to create significant interference.
That is, an AP cannot know the channel selections of other APs,
condition \nox and, moreover, cannot reliably identify or enumerate
the clauses in which it participates, condition \noM.  A decentralized
algorithm is therefore mandated.

\begin{figure}
\centering
\includegraphics[width=3in]{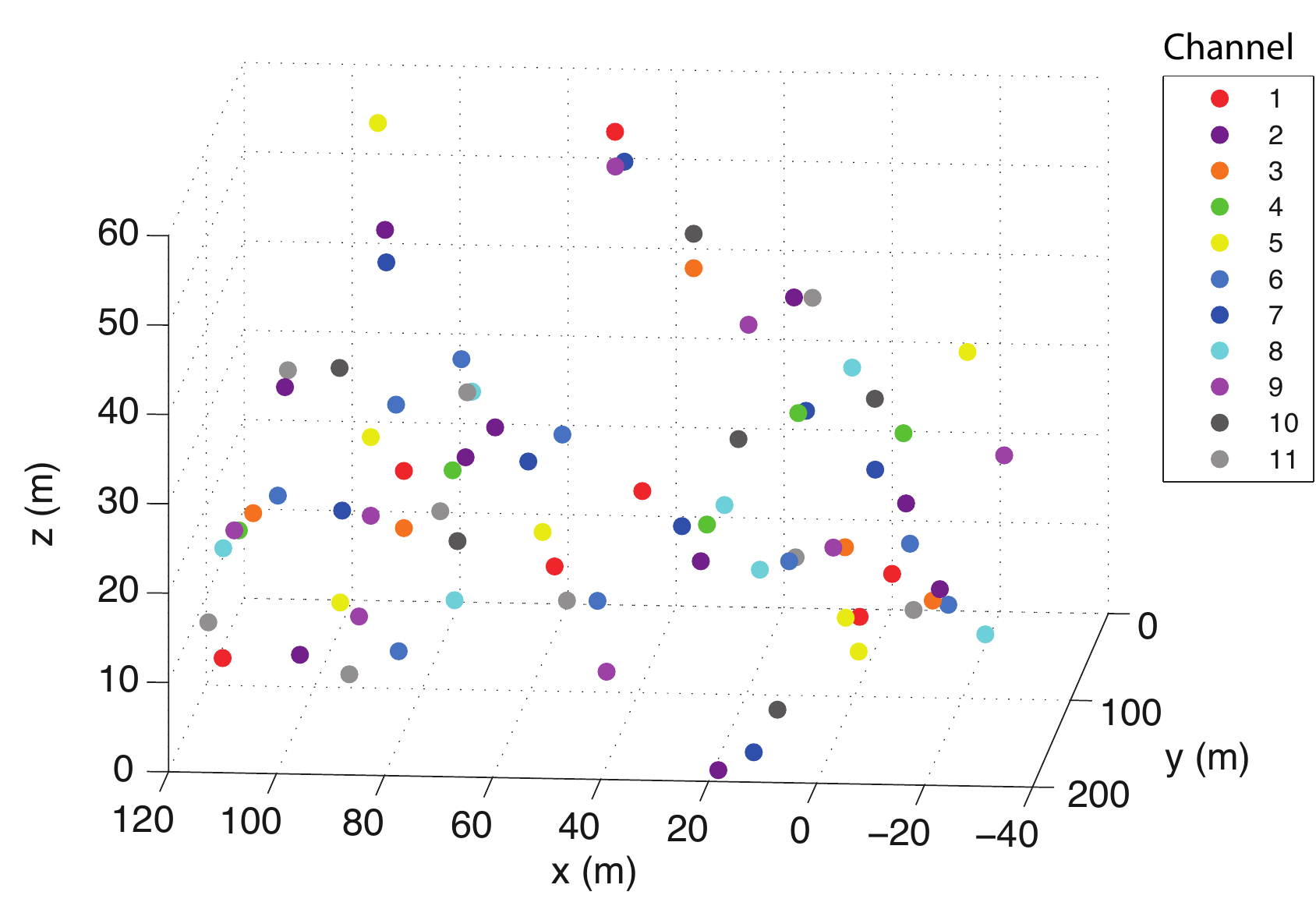}
\caption{A satisfying assignment of radio channels. Each dot marks the location of a WiFi wireless access point and is based on measurements taken at the junction of 5th Avenue/59th
Street in Manhattan. The color of a dot indicates the radio channel. To avoid interference
between transmissions, nearby access points need to operate on radio
channels that are spaced sufficiently far apart. There are 11 radio
channels available to choose from, but the frequencies of these
channels overlap so that only 3 orthogonal/non-overlapping channels
are available; there are 81 wireless access points in total.}
\label{fig:5thAve}
\end{figure}

\begin{figure}
\centering
\includegraphics[width=3in]{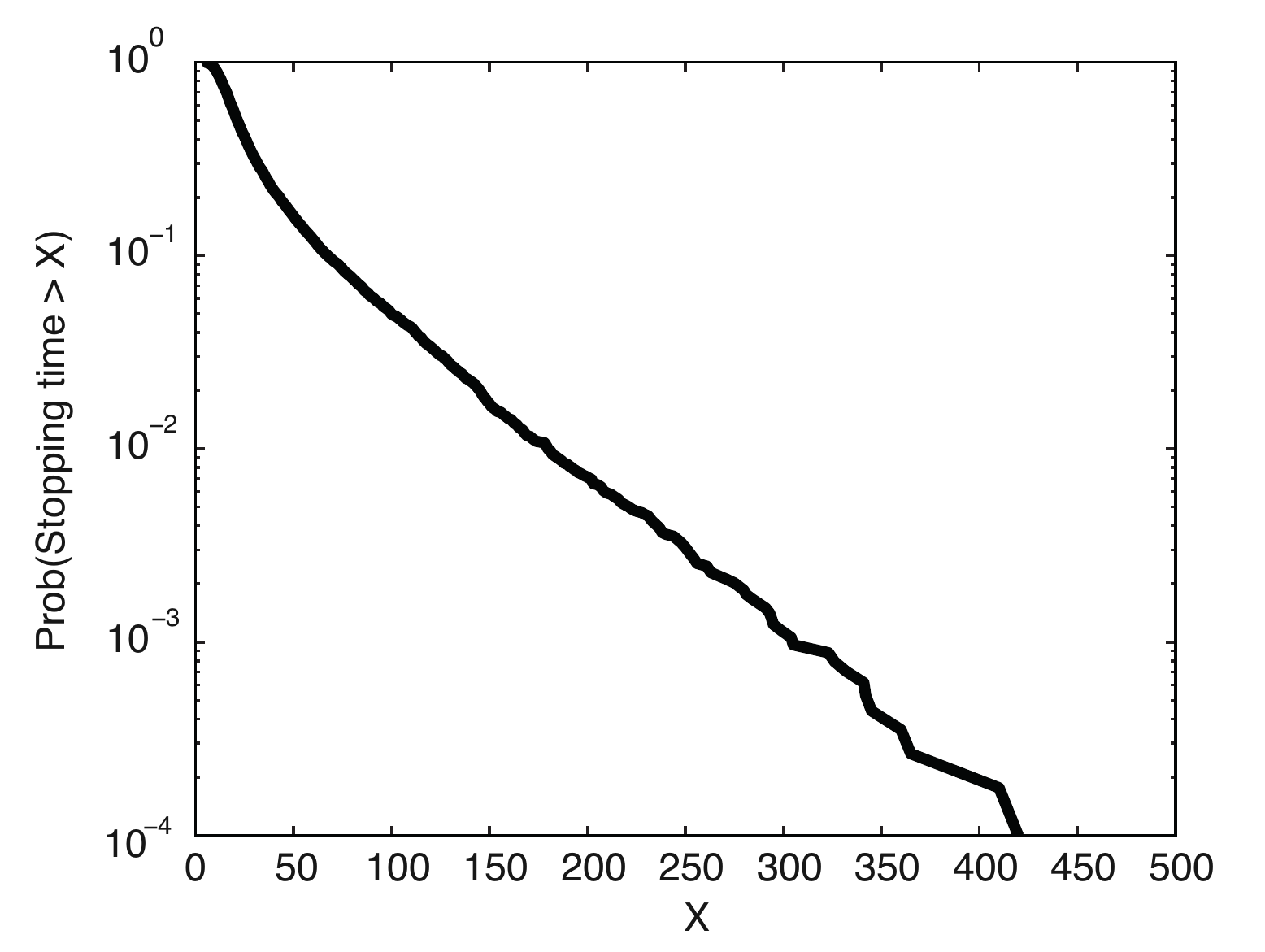}
\caption{Log of the empirical complementary cumulative distribution
of convergence time in iterations, based on 12,000 runs of CFL
algorithm for 5th Avenue data. The median is 21 and the 95\%
percentile is 98 iterations. In current hardware one iteration can
readily be performed in under 10 seconds, leading to a median time
to convergence of less than 4 minutes.}
\label{fig:5thAveDist}
\end{figure}

Fig. \ref{fig:5thAve} shows an example satisfying assignment of
radio channels obtained by running an instance of the CFL algorithm
at all APs. The complexity of the topology generated by the physical
location of the APs and the non-uniformity of the clauses it causes
is apparent.

Fig. \ref{fig:5thAveDist} shows the measured distribution of number of
iterations required to find a satisfying assignment, whereupon the
algorithm natural halts in a decentralized fashion. The median value
is 21 iterations and the 95\% centile is 98 iterations. Note that
during this convergence period, although the network is operating
sub-optimally, it does not cease to function. In a prototype lab set-up
we have shown that a CFL update interval of less than 10 seconds
is feasible on current hardware. Thus the median time to convergence
is under 4 minutes. This is a reasonable time-frame for practical
purposes, particularly as subparts of the network are functioning
during this convergence period, and thus the CFL algorithm offers
a pragmatic solution to this difficult decentralized CSP for which
existing solvers could not be employed.

\section{Conclusions}
\label{sec:disc}

We have shown that apparently distinct problems in networking can
be placed within the framework of CSPs, where - unlike with traditional
motivating examples - the variables are co-located with devices
that may not be able to communicate. We define the criteria that
practical solvers must possess in order to be suitable for theses
problems, labeling them decentralized CSP solvers.

As existing solvers fail to meet one or more of these criteria, we
introduce a decentralized algorithm for solving CSPs. We prove that
it will almost surely find a satisfying solution if one exists. In
doing so, we generate bounds for the speed of convergence of the
algorithm. We suspect, however, that our bounds for a general CSP
are not tight and conjecture that the real bound should be closer
to the one we have for the specific case of CSPs corresponding to
graph coloring.

Given how much information decentralization is sacrificing,
surprisingly an experimental investigation of solving random k-SAT
instances suggests that the algorithm is competitive with two of
the most promising centralized k-SAT solvers, WalkSAT and Survey
Propagation, on instances of random k-SAT with order one thousand
variables. This raises the question: what is the performance cost
of decentralized operation? This is particularly pertinent as the
decentralized nature of the CFL algorithm lends itself to parallelized
computation as it has a small, fixed memory requirement per variable,
making it suitable for use in circumstances where a centralized
algorithm could also be used but would be difficult to implement
in a distributed fashion.

\appendix
\section{Convergence Proof}
\label{app:proof}

For each $i\in\{1,\ldots,N\}$ let $\pv_i(t)\in[0,1]^D$,
$t\in\N$, be the CFL probability vector for variable $i$ at time
$t$ and $x_i(t)$ be the variable's value selected stochastically
from $\pv_i(t)$. Let $P(t)=(\pv_1(t),\ldots,\pv_N(t))$ and
$X(t)=(x_1(t),\ldots,x_N(t))$ record the overall state of the CSP-CFL
system.

The state of the probability vectors $\{P(t)\}$ forms a Markov
Chain, or Iterated Function System, with place dependent probabilities
\cite{barnsley88}. The convergence time of the algorithm is the
first time the chain enters an absorbing state representing a valid
solution to the CSP. The absorbing states are identifiable
in terms of the variable values $\{X(t)\}$: 
\begin{align*}
A=\bigcup\left\{\vx:\Phi_m(\vx)=1 \text{ for all } m\in \Mc\right\},
\end{align*}
as if $X(t)\in A$, then $P(t+1)=(\delta_{x_1},\ldots,\delta_{x_N})$
so that $X(t+1)=X(t)$ almost surely and hence a solution to the CSP
has been found. The algorithm's stopping time is
\begin{align*}
\tau := \inf_{t\geq0} \left\{X(t)\in A\right\}.
\end{align*}
Define $\gamma = \min(a,b)/(D-1+a/b)$ and let $S$ be the set
of states such that for all $i \in \Nc$ either $p_i=\delta_k$ for
some $k\in \{1,\cdots,D\}$ or $(p_i)_k \geq\gamma$ for all
$k\in\{1,\cdots,M\}$. The next lemma is obvious.

\begin{lemma}\label{le:uniflowerbound}
For any integer $t\geq 0$, $P(t)\in S$.
\end{lemma}

\begin{proof}[Theorem \ref{thm:main}]
The method of proof for both statements is similar: we create a
sequence of events over $N-1$ iterations that, regardless of the
initial configuration, lead to a satisfying assignment with a
probability that we find a lower bound for. Due to the Markovian
nature of the algorithm and the independence of the probability of
this event on its initial conditions, if this event does not occur
in $N-1$ iterations, it has the same probability of occurring in
the next $N-1$ iterations. This is what leads to the geometric
nature of the bounds. The difference between a general CSP and those
corresponding to coloring is that for the latter the variables 
are ensured to experience fewer dissatisfaction events before
finding a satisfying assignment.

First consider a general CSP. Select an arbitrary valid solution
$\va\in A$ with components $(\va)_i$. For each $m\in\Mc$ define
$\Nc_m$ to be the variables that participate in clause $m$.  For
each $t\geq0$, define the set of unsatisfied variables at time $t$
by
\begin{align*}
U_t :=\bigcup_{m\in \Mc} \{\Nc_m:\Phi_m(X(t))=0\}
\end{align*}
and the number of unsatisfied variables to be $n_t =|U_t|$. At
time $0$ assume $X(0)=\vx(0)$, some $\vx(0)$. If $U_0=\emptyset$ then
$X(0)\in A$, the algorithm has found a solution and $\tau=0$. If
$U_0\neq\emptyset$, the algorithm has not yet found a solution.
Define the components, $\vx(1)_i$, of the vector $\vx(1)$ by
\begin{align*}
\vx(1)_i = \begin{cases} 
	\va_i	& \text{ if } i \in U_0\\
	\vx(0)_i	& \text{ if } i \notin U_0\\
	\end{cases}
\end{align*}
and, by Lemma \ref{le:uniflowerbound},
\begin{align*}
P(X(1) = \vx(1)|X(0)=\vx(0)) \geq \gamma^{n_0}.
\end{align*}
If $U_1=\emptyset$ then
$X(1)\in A$, the algorithm has found a solution and $\tau=1$. If
$U_1\neq\emptyset$, the algorithm has not yet found a solution and
we proceed in an iterative fashion. For $t\geq0$,
define the components, $\vx(t+1)_i$, of the vector $\vx(t+1)$ by
\begin{align*}
\vx(t+1)_i = \begin{cases} 
	\va_i	& \text{ if } i \in U_t\\
	\vx(t)_i	& \text{ if } i \notin U_t\\
	\end{cases}
\end{align*}
and, by Lemma \ref{le:uniflowerbound},
\begin{align*}
P(X(t+1) = \vx(t+1)|X(t)=\vx(t)) \geq \gamma^{n_t}.
\end{align*}
Let $t^*=\inf\{t:n_t=0\}$ and note that
\begin{align*}
&
P(X(t^*) \in A) 
	\geq \\
&
P(X(t^*)=\vx(t^*),\ldots,X(0)=\vx(0)) 
	\geq 
	\gamma^{\sum_{t=0}^{t^*}n_t}.
\end{align*}
We wish, therefore, to place an upper bound on the sum in the
exponent. Observe that $t^*\leq N-1$, as one starts with at least
one unsatisfied variable at each stage we me must include at least
one new variable and, therefore, this procedure must terminate in
no more than $N-1$ steps with a valid assignment.  

In general, the $\{n_t^*\}$ sequence that maximizes this sum,
even though this sequence may not be feasible for a given CSP, is
\begin{align*}
n_0^*=1, n_1^*=2,\ldots,n_{N-1}^*=N,
\end{align*}
This occurs if we start with one unsatisfied variable and, in
changing to the valid solution, at each iteration one additional
variable has cause to be unsatisfied. On taking its new value,
the unsatisfied variable included at time $t$ triggers the failure
of $t-1$ clauses each of which contains $3$ variables: itself,
the variable included immediately previously and one of the other
variables that has already experienced a failure. This gives
\begin{align*}
\sum_{t=0}^{t^*}n_t 
	\leq \sum_{t=0}^{N}n_t^*
	= \frac{N(N+1)}{2}.
\end{align*}
and therefore
\begin{align*}
P(X(N-1) \in A) \geq 
	\gamma^{N(N+1)/2}.
\end{align*}

For a CSP corresponding to graph coloring with $D$ colours, the
advantage is that variables cannot be dissatisfied indefinitely by
newly dissatisfied variables. Instead this procedure generates a
flame-front of dissatisfied variables where those variables
sufficiently far within the interior of this flame-front can select
their final value without further disturbance. For graph colouring,
clauses occur only for each pair of neighbors, $i$ and $j$, such
that $\Phi(\vx)=0$ if $(\vx)_i=(\vx)_j$ and $\Phi(\vx)=1$ otherwise.
For any initial configuration $\vx(0)$ with dissatisfaction, we
must have $n_0\geq2$. In a sequence of events analogous to those in
the proof of \cite{duffy08}[Theorem 3], these two variables select
their final values, $\va_i$ and $(\va)_j$, at the next round, causing
their initial clause to be satisfied as $\va\in A$. If either of
the variables taking its final value causes no further clause to
be dissatisfied, then in this sequence of events it will never
appear in a dissatisfied clause again and stays with its value with
probability $1$. If it does cause one or more clause to be triggered,
once these are satisfied it will experience no more dissatisfaction.
Thus each variable experiences at most $2$ rounds of dissatisfaction
so that for graph colouring:
\begin{align*}
\sum_{t=0}^{t^*}n_t 
	\leq \sum_{t=0}^{N-1}n_t^*
	= 2N.
\end{align*}
\end{proof}

As an example that illustrates the difference between a general
CSP and graph colouring, consider an instance of 3-SAT starting
with one unsatisfied variable, $x_1$. The following sequence of
clauses is possible in general, but not with graph colouring.
Starting with an unsatisfied $x_1$, changing its value to satisfy
the first clause causes $x_2$ to be dissatisfied, which in turn
causes $x_1$ and $x_2$ and $x_3$ to be dissatisfied:
\begin{align*}
x_1,
\,\,
\neg x_1 \vee x_2
\text{  and  }
\neg x_1 \vee \neg x_2 \vee x_3.
\end{align*}
Satisfying these clauses for $x_1$, $x_2$ and $x_3$ causes two more
clauses with $x_4$ to trigger, keeping all variables unsatisfied:
\begin{align*}
\neg x_1 \vee \neg x_2 \vee x_4
\text {  and  }
\neg x_1 \vee \neg x_3 \vee x_4.
\end{align*}
Satisfying these two clauses causes three further clauses involving
$x_5$ to become unsatisfied:
\begin{align*}
\neg x_1 \vee \neg x_2 \vee x_5,
\,\,
\neg x_1 \vee \neg x_3 \vee x_5
\text {  and  }
\neg x_1 \vee \neg x_4 \vee x_5
\end{align*}
and so forth triggering clauses with three variables at a time such
that all variables are unsatisfied until a solution is found.


\end{document}